# A Multi-Layered AI Pipeline for BPMN Generation from Natural Language

**Alexandru Rancea[1], Ionut Anghel[1*], Tudor Cioara[1], Ovidiu Anchidin[2] and Ancuta Nemes[2]**

[1]Computer Science Department, Technical University of Cluj-Napoca, Memorandumului 28, 400114 Cluj-Napoca, Romania. alexandru.rancea@cs.utcluj.ro; ionut.anghel@cs.utcluj.ro; tudor.cioara@cs.utcluj.ro.
[2]Niculae Stancioiu Heart Institute, Calea Motilor 19-21, 400001 Cluj-Napoca, Romania. anchidin_ovidiu@institutulinimii.ro; nemes_ancuta@institutulinimii.ro

**Abstract:** The automated transformation of unstructured natural language process descriptions into Business Process Model and Notation (BPMN) diagrams remains a challenging task, due to linguistic ambiguity and the difficulty to detect non-adjacent control-flow relationships. This paper proposes a solution for generating BPMN models from natural language descriptions through a multi-layered pipeline that integrates transformer-based architectures, Large Language Models and ontology population. The proposed method starts with natural language descriptions and separates textual processing, control-flow relation extraction, and structural validation into dedicated components. Entity extraction is performed using a fine-tuned RoBERTa-based transformer model while a control-flow engine captures non-adjacent relationships through LLM-based inference. To improve generalization under limited available data, the approach uses a dynamic few-shot strategy with controlled template input construction, enabling improved generalization under limited data availability. Additionally, an ontology population mechanism is introduced to represent the generated process models as semantic knowledge structures. For evaluation purposes we have used from the literature and industry reference processes. Additionally, we have validated the proposed method on a real-world heart disease patients transitional care process, capturing practical constraints and multi-actor collaboration, and serving as an expert-defined benchmark scenario. Experimental results demonstrate strong performance in the relation extraction task, achieving an accuracy of 90.7%, as well as maintaining structural alignment between the generated and reference BPMN models across multiple evaluation scenarios scenarios.

## 1 Introduction

Business Process Model and Notation (BPMN) has become the default standard for representing complex organizational and industrial processes in a visual and executable form. BPMN modelled processes can be mapped to services or components allowing for orchestration or choreography of tasks thus helping automating tasks and optimizing organization specific processes [1]. In the current AI-enabled context, BPMN modelled activities can be dynamically managed by AI agents integrating learning and reasoning into the process execution [2]. However, currently BPMN model creation remains primarily a manual task, requiring domain experts to interpret unstructured natural language descriptions and translate them into structured diagrams that comply with BPMN semantics [3]. This activity is labour-intensive, subject to interpretation of errors and often leads to inconsistencies across organizations, particularly when multiple stakeholders interpret the same textual specification differently.

Automating the transformation of natural language process descriptions into formal BPMN models presents several challenges [4]. Natural language specifications usually contain implicit dependencies, ambiguous sequencing, and branching conditions that must be translated into specific BPMN control-flow structures such as gateways and parallel flows. Conventional relation extraction techniques usually assume that relationships occur between adjacent entities in text, an assumption referred to as the adjacency constraint [5]. Nevertheless, business process descriptions often include non-adjacent (i.e., non-linear) control-flow relationships, such as conditional branching and parallel execution, where related elements may be separated by multiple clauses or sentences. As a result, systems that rely primarily on sequential text interpretation can fail to accurately reconstruct the correct control-flow structure of complex processes.

Existing work has explored multiple approaches to address these challenges. Process algebra has been used for modelling and verification of business models, allowing for reasoning and checking the correctness and early identification of any logical faults in the model design phase [6]. Early attempts for rule-based natural language processing (NLP) pipelines combined syntactic parsing and semantic role labelling to identify activities and their ordering relationships [7]. Later studies extended this direction toward declarative process modelling by extracting activities and constraint relations from text and mapping them to declarative process templates [8]. The development and evolution of transformer-based architectures such as BERT [9] and RoBERTa [10] have improved token-level perception tasks, including entity recognition, coreference resolution, and relation extraction, enabling more reliable detection of process elements from textual descriptions. Furthermore, large language models (LLMs) and foundational models are lately investigated as a solution to process modelling and BPMN generation tasks [11,12]. While these models demonstrate that they can generate syntactically valid process representations, empirical studies indicate that the overall structural correctness tends to deteriorate as complexity increases, particularly in the presence of branching logic and non-adjacent dependencies [13]. For defining knowledge that can be reused and integrated in such BPMN generation techniques, especially in the modern AI-based ones, existing BPMN ontology semantically represents elements, relationships, and rules [14]. To add real execution context and transforming this knowledge base from a static model into a representation of how processes behave requires expanding it with process instances. This would allow Artificial Intelligence (AI) models such as LLMs to better learn and improve their ability to generate context-aware insights. These findings highlight the need for further research and development of novel architectures that combine modern AI techniques with semantic knowledge representation and explicit structural validation mechanisms capable of generating valid BPMN processes that can be further used for automating domain specific processes and even integrating AI agents into the process flow [15].

In this context, we propose a multi-layered approach for transforming unstructured natural language descriptions into valid BPMN models. The main contributions of this work are:

- Design of a multi-layered pipeline for automated BPMN generation from natural language descriptions while separating entity identification, control-flow extraction, and structural validation into dedicated processing layers.
- Definition of an entity identification method that combines Longformer architecture and a fine-tuned RoBERTa-based transformer model to classify tokens into BPMN primitives.
- Definition of a relation extraction engine that combines structured candidate generation with few-shot LLM-based inferences to capture non-adjacent (non-linear) control-flow relationships in the BPMN process descriptions.
- Definition of a validation layer that assures that the resulting BPMN structure complies with the BPMN modelling constraints before generating the final representation. It also integrates an ontology population mechanism that extends the existing BPMN ontology with a semantic knowledge representation of the generated process model, enabling semantic persistence.
- Evaluation and validation of the proposed method on expert defined process models of real-world scenarios and on literature defined process models.

The rest of this paper is organized as follows: Section 2 reviews the state of the art in automated BPMN generation from natural-language descriptions, categorizing existing approaches according to their methodological approaches and highlighting the current research gaps. Section 3 details the proposed method. Section 4 presents evaluation and validation results including component level and overall performance analysis through multiple case studies. Section 5 discusses the method results, analysing the strengths and limitations and their applicability in real world scenarios. Finally, Section 6 concludes the paper by summarizing the main contributions and highlighting directions for future research.

## 2 Related Work

The automated transformation of natural language process descriptions into BPMN diagrams has been thoroughly studied, driven by the need to accelerate process digitalization and reduce the specialized knowledge required for manual modelling. Existing approaches in the literature can be grouped into five categories: rule-based NLP pipelines, declarative or constraint-based extraction methods, machine learning (ML) and data-driven extraction approaches, modern LLM-based generation frameworks, and structured intermediate-representation approaches. In the following we analyse these categories and discuss their contributions and limitations.

Early contributions focus on rule-based NLP pipelines that mapped linguistic structures to imperative modelling notations. [16] combined syntactic parsing, semantic role labelling, and a dedicated anaphora resolution to extract activities and ordering relations from textual specifications. Their approach constructed an intermediate "world model" before generation of BPMN diagrams, demonstrating the technical feasibility of systematic text-to-model transformation. Similar approaches remain relevant for domains where explainability and deterministic behaviour are required, as argued in [17] and [18]. However, rule-based solutions require extensive linguistic rules and domain knowledge, making it difficult to adapt to new domains or complex process descriptions. Moreover, such approaches fail to capture implicit relations and non-adjacent control-flow structures that are usually found in natural language specifications.

Subsequent research extended process extraction beyond imperative control-flow reconstruction toward semantically richer and more flexible representations. [19] introduced the first automated approach for extracting declarative process models from natural language. Instead of reconstructing explicit sequences, their approach identifies activities and constraint relations and maps them to Declare templates. This shift from sequence-oriented modelling to constraint-based extraction reflects a broader recognition that textual description often specifies constraints and obligations rather than enumerated execution sequences, as described in [20]. Declarative approaches focus on constraint identification and do not directly construct structures required for BPMN diagrams, and consequently additional transformation steps are often required for process representations. In parallel, research has addressed the misalignment that frequently occurs between textual description and process models. Techniques for automatic inconsistency detection between text and process diagrams have demonstrated discrepancy in ordering, missing activities and constraint violation are common occurrences, as presented in [7]. Furthermore, compliance checking approaches based on behavioural spaces explicitly model ambiguity by representing multiple plausible interpretations of natural language specifications instead of merging into a single deterministic structure, as discussed in [8].

Methodological approaches to process extraction have evolved significantly with the increasing adoption of ML and deep learning techniques in NLP. [21] documents the transition from rule-based pipelines toward hybrid and learning-based architectures. Recent studies address this transition by decomposing text to process modelling into standard information extraction subtasks, such as named entity recognition and entity linking, by implementing these components with trainable models rather than handcrafted linguistic rules. [22] formulates process model extraction as a text classification problem and uses hierarchical neural architecture to capture process relevant details across different textual levels. [23] replaces rule based named entity recognition and relation extraction with fully data-driven pipelines, enabling faster adaptation across datasets, types and domains. Furthermore, structure aware NLP approaches, as described in [24], have been proposed that use flexible tree structured patterns over dependency parses to extract actions, roles, conditions, and control-flow relations, offering generalization to linguistic variation, while requiring less rule management compared to earlier handcrafted methods. While these methods show improved validity in linguistic analysis; their effectiveness remains affected by the limited annotated datasets for processing extraction tasks. The increasing use of such methods highlighted the need for standardized datasets and evaluation benchmarks for processing extraction tasks. One of the first benchmarks of text-to-process extraction techniques is defined in [25] and highlighted the lack of annotated datasets as a major limitation in this field. The introduction of PET in [26] represents a major step toward standardized evaluation, providing annotated descriptions that include activities, actors, gateways, and flow relations. Even so, dataset limitations remain one of the primary bottlenecks for such approaches.

The recent evolution of transformer-based and large language models (LLMs) has reshaped research directions in business process management (BPM). [27] systematically analyses the integration of LLMs across the BPM lifecycle, identifying opportunities in modelling support, documenting, and conversational interaction, while also showcasing methodological risks related to hallucinations, evaluation and reliability. In the context of process modelling, [11] propose structured approaches for LLM-based model generation that incorporates abstraction choices and iterative improvement strategies. Furthermore, [12] presents an evaluation framework and benchmark, for assessing LLM performance on business process modelling tasks. Their findings show deteriorates LLMs can generate syntactically valid models, but the structural correctness and the overall quality deteriorate as complexity increases. This is a finding that is further supported by [13], which evaluated BPMN XML diagram generation for robotic process automation scenarios across different complexity levels. It indicates strong performance for basic processes but decreasing structural correctness and validity for more advanced processes that involve complex gateways. These results suggest that basic one-shot diagram generation is insufficient for process modelling, and additional mechanisms are necessary. The usage of LLMs in the context of BPM can tackle the issue of process model comprehension. [28] shows that the abstraction choice directly influences LLM performance in understanding and reasoning over process models. Representations that expose structural semantics in a machine interpretable format enhance their reasoning capabilities compared to graphical encodings. This suggests that LLMs can not only be used as generators in the area, but as interpreters that support validation, querying, and iterative correction within process modelling. Similarly, [4] introduces BPMNGen, a conversation framework that generates BPMN 2.0 process models from natural language descriptions using a custom LLM assistant. The system produces intermediate JSON representations that are later converted and validated as BPMN XML diagrams. While such approaches demonstrate the feasibility of text-to-BPMN generation, it highlights as well that many similar applications remain in conceptual prototypes due to the practical complexity of such a system. Looking at research outside the BPMN context, process generation in the LLM literature provides relevant methodological insights. System such those presented in [22] and [23] use LLMs to generate executable processes using APIs and stages of prompting strategies. Hallucinations are addressed by applying validation constraints within the predefined function interfaces and including user feedback for iterative improvement. They demonstrate that decomposition and controlled abstraction successfully improve the generation's validity.

Another focus of recent research is the use of intermediate representations (IRs) to improve the controllability and reliability of model generation. BPMN Assistant, presented by [29], demonstrates that generating and editing JSON representations before compiling them into BPMN XML improves edit success rates while maintaining comparable similarity to ground-truth models. By moving away from the verbose XML syntax, JSON-based IRs reduce the cognitive and structural overhead position on LLMs and enable validation before the final export. Similarly, the Partially Ordered Workflow Language (POWL), as shown in [30], introduces a modelling formalism that integrates partial orders with hierarchical control-flow operators. By design, POWL provides soundness guarantees and offers a compact representation of concurrency, choice, and loops. Although not designed specifically for LLM-based pipelines, such representations are well suited as intermediate stages in the generation step. However, the intermediate representations alone do not guarantee semantic equivalence between the general workflow and the natural language description. Although structured model-to-text generation has been demonstrated in [31], current LLM-based text to BPMN systems rarely incorporate reverse verbalization as a validation step. Such an approach, following a text-to-model generation, could provide a useful mechanism for verifying semantic equivalence between the original specification and the generated process, such as detecting ambiguities or structural inconsistencies.

Overall, the current state of the art reflects a transition from rule-based text-to-model transformation toward a multi-stage generation pipeline that combines semantic extraction, constraint modelling, intermediate representations, and iterative improvement. While LLMs expand the potential for automation in process modelling, they still require extensive validation layers and evaluation frameworks to ensure reliability. The research is moving beyond whether the process models can be generated from text, towards how generation can be validated, controlled, and integrated in complex process models.

## 3 Materials and Methods

The proposed architecture is a multi-layered transformation pipeline designed to convert natural language specifications into a valid BPMN 2.0 representation. The methodology uses a set of three primary architectural layers (Entity Identification Layer, Relation Extraction Layer and Validation Layer) across multiple stages, as illustrated in Figure 1.

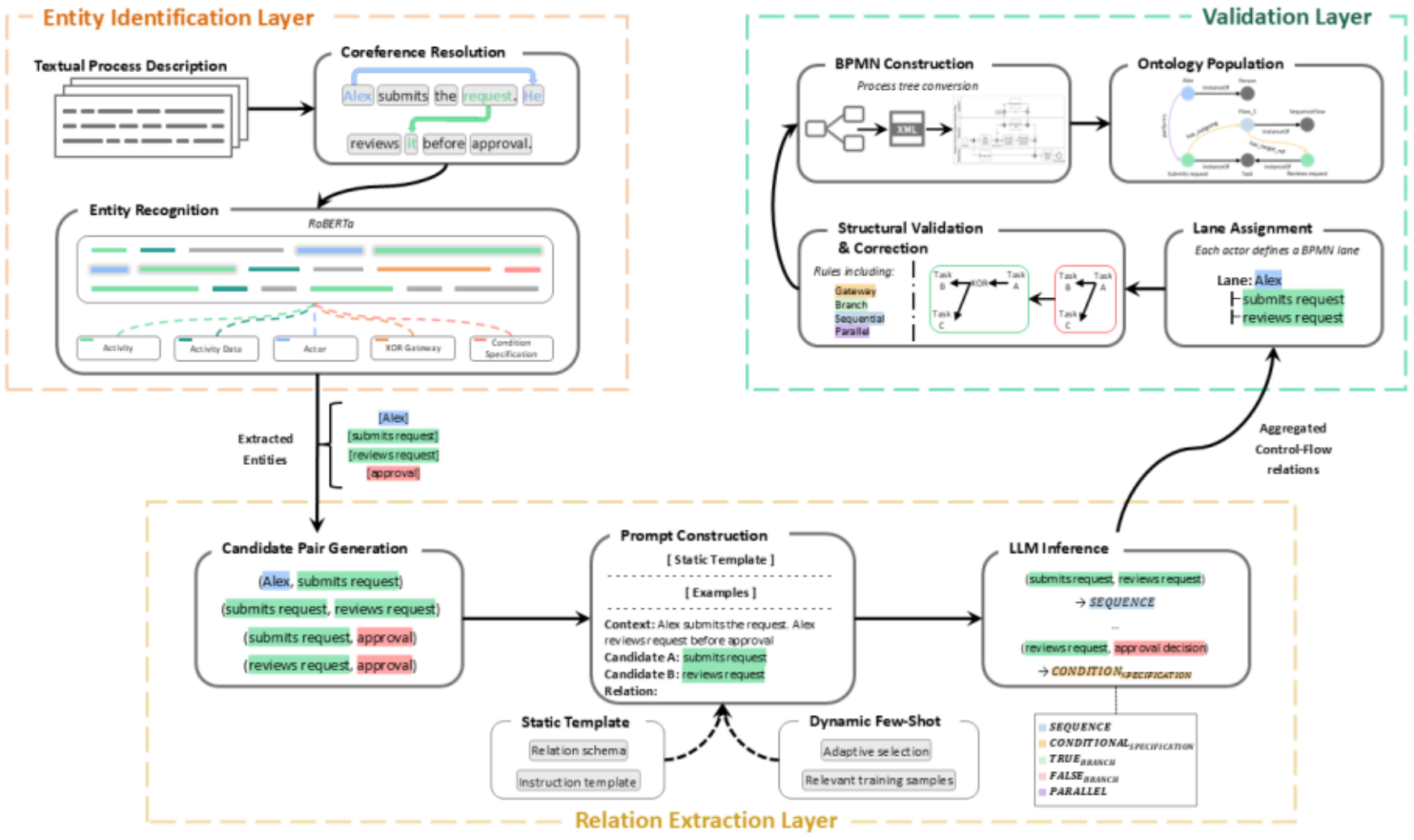

**Fig. 1** Architectural overview of the proposed method.

Formally, given an unstructured natural language process description, the objective of the proposed pipeline is to generate a BPMN process model $M$, that can be formally defined as:

$$M = (E, R) \quad (1)$$

where E represents entities and R relations.

$$E = \{e_{\mathrm{i}} \mid \mathrm{i} \in \{1, 2, \dots, \mathrm{n}\}\} \quad (2)$$

$$R = \{\mathrm{r}_{\mathrm{ij}} \mid e_i, e_j \in \mathrm{E}\} \quad (3)$$

As defined in Equation (2), the entity set includes all the extracted process entities. Equation (3) defines the set of control-flow relationships between entities. Each relation $r_{ij}$ represents a direct relationship between entity $e_i$ to entity $e_j$, capturing the order, branching structure or parallel flow.

Entities are further categorized into three primary BPMN classes: activities, gateways, and actors.

$$E = A \cup G \cup U \quad (4)$$

$$\mathrm{A} = \{a_{\mathrm{i}} \mid \mathrm{i} \in \{1, 2, \dots, \mathrm{k}\}\} \quad (5)$$

$$G = \{g_i \mid i \in \{1, 2, \ldots, l\}\} \quad (6)$$

$$U = \{u_i \mid i \in \{1, 2, \ldots, m\}\} \quad (7)$$

where A denotes the activities representing executable tasks or actions within the process, G denotes control-flow elements that define branching, merging, or parallel execution and U represents the participants or entities responsible for executing the activities.

The generation process begins with the Entity Identification Layer, a preprocessing step that resolves pronominal references through coreference resolution to ensure semantic completeness of textual specification. This normalized input is processed by a fine-tuned transformer-based entity recognition module, which classifies tokens into BPMN primitives. The extracted entities are passed to the Relation Extraction Layer, that combines candidate generation with a LLM engine to infer dependencies between process elements. Finally, the inferred structure is processed within the Validation Layer where entities are assigned to BPMN lanes based on identified actors, then structural validation constraints based on BPMN ontology schema are used to detect and correct inconsistencies in the generated control-flow graph. Once validated, the model is serialized into a BPMN 2.0 XML format and the extracted entities and their relationships are instantiated as ontological individuals within the BPMN ontology, enabling semantic knowledge representation and future enhancement of LLM-based inference.

## *3.1 Entity Identification Layer*

The Entity Identification Layer (see Figure 2) aims to identify BPMN-relevant elements that serve as the fundamental components of a process model. Process descriptions frequently contain pronominal references (e.g. He, She, etc.) that hide the association between action and the actors performing them. The standard Named Entity Recognition (NER) models often process tokens independently and may fail to associate correctly with the appropriate process participants. To mitigate this, we have integrated a preprocessing module based on FastCoref, as defined in [32], a coreference resolution model based on the Longformer architecture, [33]. The module scans the input text, identifies the coreference clusters and replaces all the pronominal mentions with their referents. It converts ambiguous phrases such as "He validates it" into explicit statements as "The manager validates the invoice". This transformation produces a semantically explicit representation of the description, which improves the reliability of the extraction and actor identification tasks.

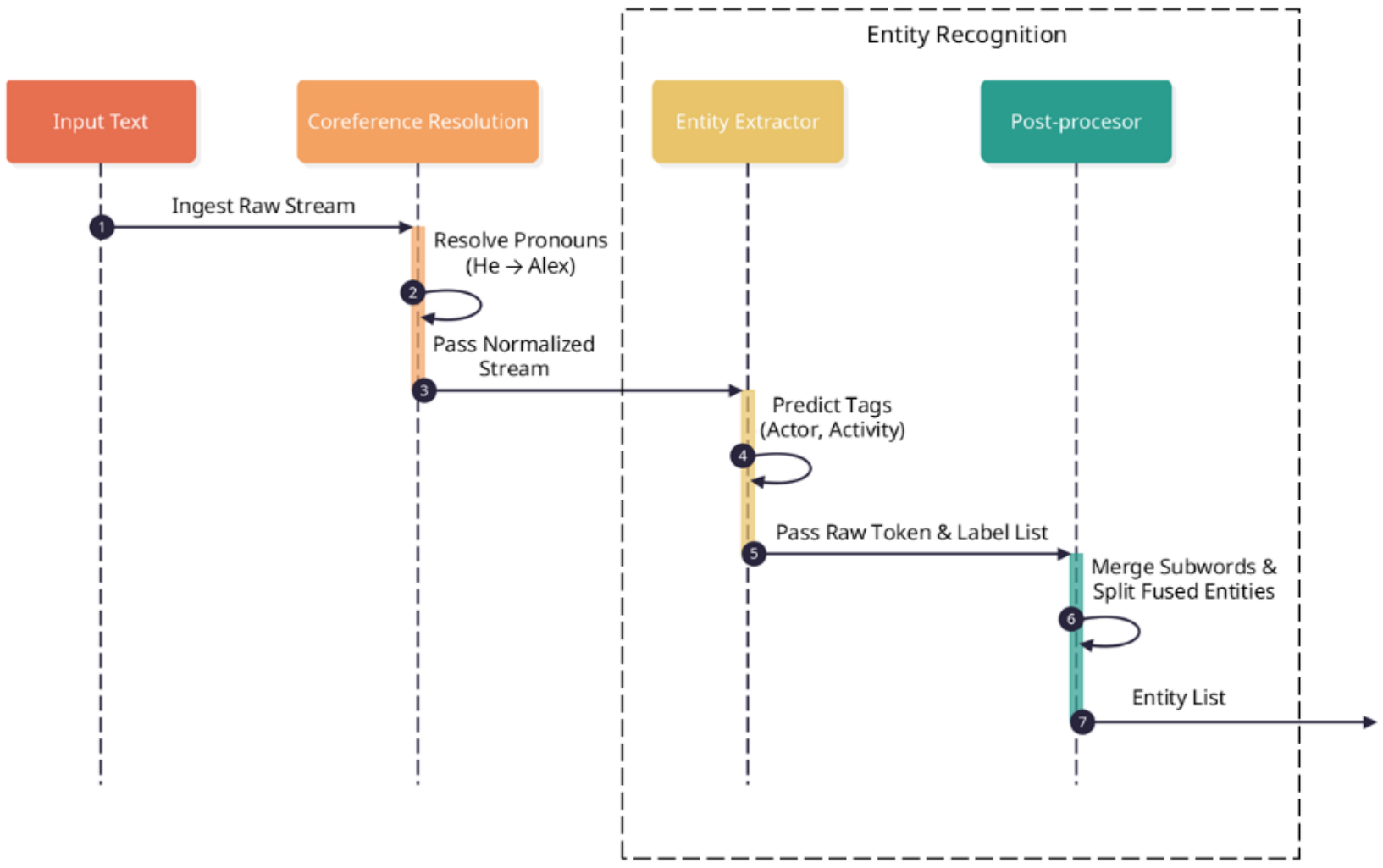

**Fig. 2** Architectural workflow of the Entity Identification Layer.

For Entity Recognition, first **Entity extraction** is performed using a fine-tuned RoBERTa-based transformer model [10]. The model operates as a token classifier using the BIO (Beginning-Inside-Outside) tagging scheme. Given input sequence, the model predicts a corresponding sequence of elements each belonging to the entity tag set corresponding to the BPMN elements. In addition to identifying the main entity classes, the extraction process may capture auxiliary contextual information associated with the activity descriptions, such as contextual attributes. This information is preserved as a metadata linked to the corresponding activity entity and can be utilized by other layers of the pipeline. Then a **post-processor** component is used to rectify common prediction errors. While Transformer architecture provides superior semantic mapping, they often lack a strict schema format required for further layers. We bridge this gap by implementing two deterministic correction functions:

- Subword Merging: Transformer tokenizers frequently split words into subword fragments. A reconstruction procedure scans the predicted token sequence and merges contiguous subword tokens that share the same entity label, restoring the complete lexical units.
- Entity Splitting: In certain cases, the model incorrectly merges gateways indicators with adjacent activity descriptions. A rule-based separation procedure identifies predefined BPMN logic markers and separates them from surrounding tokens, ensuring that gateway expressions are represented as independent entities.

The output of the Entity Identification Layer is the entity set E, which is passed to the Relation Extraction Layer, where control-flow relationships between the extracted entities are inferred.

## *3.2 Relation Extraction Layer*

Figure 3 presents the Relation Extraction Layer, showing the candidate generation and inference process.

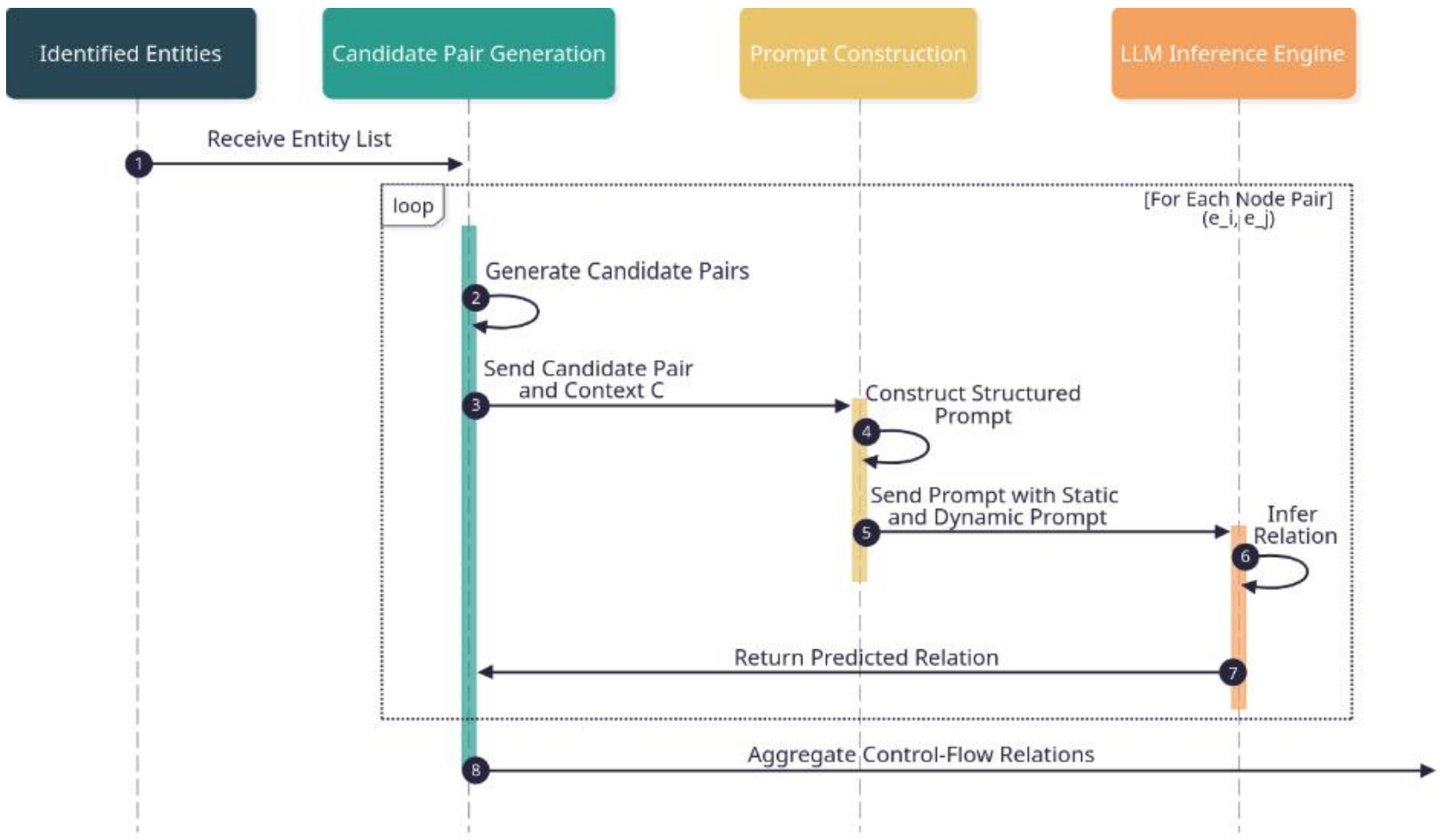

**Fig. 3** Architectural workflow of the Relation Extraction Layer.

This module starts with a multi-class relation extraction problem. Given the ordered entity sequence, as presented in Equation (2), the model predicts a relation label:

$$r_{ij} = T(e_i, e_j) \tag{8}$$

The label space used for classification is formally defined in Equation (9).

$$T = \left\{ \begin{matrix} SEQUENCE, TRUE_{BRANCH}, FALSE_{BRANCH}, \\ CONDITION_{SPECIFICATION}, PARALLEL \end{matrix} \right\} \tag{9}$$

where:

- $SEQUENCE$ represents the default sequential flow between activities
- $TRUE_{BRANCH}$ represents the positive conditional branch of a gateway
- $FALSE_{BRANCH}$ represents the negative or exception branch
- $CONDITION_{SPECIFICATION}$ captures the association between a gateway and its condition clause
- $PARALLEL$ represents concurrent execution paths

Natural language process descriptions often contain non-adjacent control-flow dependencies, where related entities are separated by multiple clauses or sentences, especially in conditional and branching structures involving gateways. To address this, candidate relations are generated using a sliding window mechanism over the entity sequence. For each entity $e_i$ the system evaluates relations within a look-ahead window of size k:

$$P = \{(e_i, e_j) \mid j - i \leq k\} \tag{10}$$

In this work, the window size is set to k = 3, which balances the detection of non-adjacent relations against computational complexity. This mechanism enables the model to evaluate both proximal and non-adjacent relations that are common in conditional process structures.

The candidate relations are evaluated by the **candidate pair generation**, which performs contextual analysis to determine the most possible control-flow dependency between the entities. Instead of relying only on adjacent dependencies, the engine analyses linguistic clues such as temporal ordering, logical connectors, and conditional expressions in the surrounding text. The context C represents a textual segment extracted from the normalized process description, containing $e_i$ and $e_j$ along with their surrounding context, such that it captures the semantic information required to infer their control-flow relationship. To represent the relation extraction task, each candidate entity pair, together with its context C, is encoded using the input.

The input is processed by a **LLM inference engine** [34] within the control-flow engine to predict a relationship between the entities. It was selected due to its parameter efficiency, large context window, and its ability to perform chain-of-thought reasoning. Model interactions are managed using the DSPy framework, as described in [35], which enables prompt construction and optimization. Given a triplet ($e_i$, $e_j$, C) consisting of the candidate entity pair and the surrounding textual context C, the model predicts the most plausible relation label $r_{ij}$ while simultaneously generating an intermediate tracing. This process allows the system to infer implicit logical dependencies that may not be explicitly expressed by syntactic patterns. To overcome the limitations of the zero-shot inference, the relation extraction component uses a dynamic few-shot conditioning strategy that combines both static and dynamic prompt construction. The static component consists of a fixed template that defines the relation classes, the input structure, and the expected output format. This includes the task description, a list of rules describing each relation type, and a schema specifying the context, candidate entity pair, and predicted relationship. The dynamic component is based on programmatic selection of annotated examples from the training data. Each sample contains the contextual sentence, the candidate entity pair ($e_i$, $e_j$) and the corresponding relation label. During inference, a subset of three of these examples is selected using a relevance sampling strategy and included in the input sequence.

**Few-Shot Prompt example**

Analyse the logical relationship between two process steps (Node A and Node B) based on the context.
Relation types:
1. SEQUENCE: Node B follows Node A chronologically.
2. CONDITION_SPECIFICATION: Connects a gateway or conditional marker to the condition.
3. TRUE_BRANCH: Action executed when the condition is satisfied.
4. FALSE_BRANCH: Action executed when the condition is not satisfied.
5. PARALLEL: Activities executed concurrently.

Output format:
Context: <process fragment>
Node A: <source node>
Node B: <target node>
Relation: <one of [SEQUENCE, TRUE_BRANCH, FALSE_BRANCH, PARALLEL, CONDITION_SPECIFICATION]>

Example 1:
Context: If the payment fails, the system sends an alert. Otherwise, it proceeds to shipping.
Node A: Otherwise
Node B: proceeds to shipping
Relation: FALSE_BRANCH
…

Context: Once both the loan risk assessment and the property appraisal are complete, the Loan Officer assesses the applicant's eligibility. If the applicant is eligible, the Loan Officer prepares the acceptance pack. Otherwise, the loan officer rejects the application.
Node A: Otherwise
Node B: rejects the application
Relation:

For more complex branching logic, we have adapted the BoostrapFewShot optimizer, as presented in [35], by defining a validation step that penalizes the model if it fails to generate the exact BPMN relation class for given description. With this step-in place, the model iteratively evaluates model predictions and selects demonstrations that maximize classification accuracy against the training set. This approach allows the generated demonstrations to guide the model to distinguish between nested logical pathways without relying on manually designed heuristics.

The predicted relations are aggregated into a structured representation describing the control-flow dependencies between the extracted entities. This representation captures the ordering and branching relationships inferred from the textual description of the process and serves as the input for the Validation Layer, where BPMN structural constraints are applied and the final BPMN model is generated.

### 3.3 Validation Layer

While the previous stages identified process entities and their control-flow dependencies, this one ensures that the resulting structure complies with the BPMN modelling constraints before producing the final representation. This layer performs four main actions: (i) assigning activities to BPMN lanes based on the identified actors, (ii) validating the structure of inferred control-flow relations, (iii) serializing the validated structure into a BPMN 2.0 XML file, and (iv) populating a semantic ontology representation of the generated process model.

Actors identified in the Entity Identification Layer define the organizational structure of the generated BPMN model. Each **identified actor** (as defined in Equation (7)) corresponds to a BPMN lane, representing the participant responsible for executing a subset of activities. Each activity is mapped to the lane associated with its corresponding actor. Activities sharing the same actor are grouped within the same lane, ensuring consistency between the textual process description and the generated BPMN model.

Before generating the final BPMN model, the **inferred control-flow relations are validated** against a set of structural constraints derived from the BPMN specification. These ensure that the generated process is syntactically correct and executable. The validation procedure verifies several structural properties:

- Gateway correctness, ensuring that conditional branches originate from gateway elements.
- Branch completeness, ensuring that gateway nodes contain the required outgoing flows.
- Sequential consistency, ensuring that activities participate in valid incoming and outgoing relations.
- Parallel flow integrity, ensuring that concurrent branches are correctly defined.

When inconsistencies are detected, rule-based validations are applied to correct the structure before serialization. This validation step prevents the generation of invalid BPMN models and ensures that the resulting process diagrams follow the BPMN control-flow semantics.

Following structural validation, the system **serializes the validated BPMN** into standard BPMN 2.0 representation. As an intermediary step, the extracted entities and relations are first converted into a Process Tree representation using the PM4Py library [36]. Compared to direct BPMN generation, this method ensures that the resulting structure consists of properly nested blocks free of deadlocks. The tree is then serialized into a baseline BPMN XML structure. The resulting BPMN model is therefore structurally valid, semantically enriched and compliant with the BPMN 2.0 specification, enabling direct visualization, further editing, or execution within the standard BPMN process engines.

Finally, a **semantic ontology representation** of the generated process model is created using Web Ontology Language (OWL). The extracted entities and inferred relations are instantiated as ontological individuals within the BPMN ontology. This process creates a semantic representation of the generated process in the form of a knowledge graph. Each entity $e_i \epsilon E$ is mapped to a corresponding ontology class instance, while each relation $r_{ij} \epsilon R$ is represented as an object propriety linking the corresponding entities. This semantic representation preserves the structural and organizational relationship identified during the extraction process. Maintaining the process models generated within an ontology enables future semantic querying of process structures, facilitates knowledge reuse across related processes, and allows future mechanisms to refine or extend previously generated. We have used the Python framework owlready2 [37] that creates a copy of the original a previous artifact based on BPMN 2.0 Ontology OWL file format, which contains the definitions of the meta-model classes and their associated properties. For each identified entity, a corresponding OWL Individual is created, mapping it to its ontological class, and the entity's attributes (such as name, lane assignment) are injected as Data Properties. When there is a relation between two entities, firstly the system updates the *has_outgoing* property of the source, to point to a new *SequenceFlow* instance, which is linked to the successor entity using the *has_target_ref* property.

# 4 Evaluation results

## *4.1 RoBERTa-based entity recognition model training and evaluation*

For training and evaluating the Entity Recognition component we have used the process extraction from Text (PET) benchmarking dataset [26], a collection of business process descriptions annotated for entity extraction. It is composed of annotations at the token level, using the BIO Tagging scheme [38]. Within the PET dataset, the annotated token labels are mapped to the entity categories (activities, gateways, and actors). Additional annotations, including Activity Data and Further Specification, are preserved as metadata associated with the corresponding activity. The PET dataset is composed of 45 annotated documents including various domains such as Recruitment, Energy Distribution, and Claims Handling. In total, it contains 417 sentences, which were processed into 675 logical relation pairs. The average sentence length is 18.15 tokens, with a maximum of 51 tokens necessitating models capable of capturing medium-range dependencies. The dataset includes 1,679 annotated entities, with a distribution skewed toward Actor and Activity types. AND Gateways account for only 0.5% of the total annotations, introducing a class imbalance that poses additional challenges for relation classification and model generalization (presented in Table 1).

**Table 1** Entity Distribution

| *Entity Type* | *Count* | *%* | *Description* |
|---|---|---|---|
| Activity | 502 | 29.9 | The primary actions or steps. |
| Activity Data | 459 | 27.3 | Documents, emails, or objects are processed. |
| Actor | 449 | 26.7 | People, roles, or systems performing actions. |
| XOR Gateway | 117 | 7.0 | Exclusive decision points. |
| Condition Specification | 80 | 4.8 | Specific constraints define gateways paths. |
| Further Specification | 64 | 3.8 | Additional details about the activity. |
| AND Gateway | 8 | 0.5 | Parallel execution flows. |

The model is trained to maximize the conditional probability by minimizing the Cross-Entropy Loss using the annotated dataset. A learning rate of $2 \times 10^{-5}$ as it is a common choice for fine-tuning transformer-based models and provides stable convergence without affecting pretrained representations. A linear learning rate scheduler was applied to reduce consistently the learning rate during training, improving the convergence stability. A weight decay coefficient of 0.01 was used as a regularization mechanism to constrain parameter magnitudes and reduce overfitting given the limited dataset available. The batch size of 16 was chosen as a trade-off between computational efficiency and stability, while the number of epochs was determined experimentally, with convergence seen after approximately 10 epochs. These hyperparameters were selected following standard practices for such architectures and adjusted through experiments.

For training the RoBERTa model, we have used an 85/15 training validation split using a fixed random seed to ensure reproducibility and consistent comparisons across experimental tests. This split was selected due to the tradeoff between providing sufficient training data for model learning while keeping a representative validation set to perform evaluation. All text inputs were pre-normalized using the RoBERTa Byte-Level BPE tokenizer to handle domain specific terminology and ensure tokenization across different process descriptions. To address the number of validation samples, we extended the dataset with 20 additional process descriptions designed to reflect the structural and linguistic characteristics of the PET dataset. These additional process descriptions were created using a hybrid approach; some of them were derived by adapting existing PET examples to different domains while keeping their process structure. The other descriptions have been generated using Gemini [39] following the annotation format and style of the PET dataset. This additional data improved overall the class distribution, and linguistic patterns. Before training, all textual descriptions were processed using coreference resolution to make sure that pronominal references of the same actor were resolved before the identification task.

To evaluate the effect of linguistic preprocessing, we conducted an analysis comparing performance on original PET dataset against the version in which coreference resolution was applied as a preprocessing step, summarized in Table 2. The introduction of coreference provides a substantial improvement in both accuracy and F1-score, increasing the overall accuracy with 3.5% and F1 with 12.7%. This indicates that resolving pronominal references significantly reduces entity ambiguity, most notably in Actor-Activity associations, and improves the model's ability to correctly identify minority classes. These results show the importance of integrating the coreference resolution as a preprocessing step.

**Table** 2 Named Entity Recognition Evaluation

| *Dataset Variant* | *Accuracy (%)* | *F1 Score* | *Evaluation Loss* |
|---|---|---|---|
| Original | 90.1 | 0.77 | 0.44 |
| Coref-Resolved | 93.6 | 0.89 | 0.39 |

## *4.2 LLM-based relation extraction inference engine evaluation*

For the relation extraction task, no publicly available dataset exists that maps the logical dependencies required for BPMN construction. Conventional relation extraction models are typically trained by adjacent entity pairs which assume that semantic dependencies occur sequentially in the text. However, BPMN process descriptions usually contain non-adjacent control-flow dependencies, most of the time in gateway constructs. In such cases, the relation between a gateway and its corresponding branches may be across multiple clauses. These structures often follow a pattern such as true branches are proximal relations, while false branches are usually non-adjacent control-flow dependencies. To address this, we designed a generation pipeline focused on capturing conditional branching patterns. We initiated the process by extracting sentence structures that contain explicit control-flow logic from the PET dataset. A sliding window algorithm was then applied over the extracted sentences. While traditional methods assume strict adjacency, this assumption was relaxed in our method because gateway constructs frequently separate the positive and negative branched by intermediate clauses. Since these extractor pairs lacked explicit logical labels, we used a Zero-Shot BART model [40], to automatically annotate the 519 simplified relation pairs. The model was prompted with two hypotheses: Hypothesis A: "The step {target} represents a positive flow" and Hypothesis B: "The step {target} represents a negative or rejection flow" to distinguish between true and false branches and annotate the dataset.

To further ensure generalization, an evaluation dataset was constructed to include variations in domain terminology, phrasing, and control-flow structures that are not present in the training data. This design prevents the model from relying on memorized patterns and instead evaluates the ability to infer relationships in previously unseen scenarios. Each data instance follows the same representation introduced in the methodology, consisting of the contextual sequence C, the candidate entity pair $(e_i, e_j)$, and their corresponding relation label $r_{ij}$. The dataset was created using a hybrid-generation strategy. The initial samples were hand-crafted and well-known examples that were not included in the PET dataset, with the focus on gateways and implicit negation patterns. Additional samples were generated from existing PET scenarios by altering domain-specific terminology and keywords used in the gateway components. Moreover, we have continued to generate more data based on these two methodologies. The resulting benchmark is explicitly balanced for the branching logic, ensuring that the evaluation includes sufficient examples of all control-flow dependencies. The dataset contains: 142 Branching samples, 66 Sequences, 48 Condition Specification, and 46 Parallel constructs (illustrated in Figure 4).

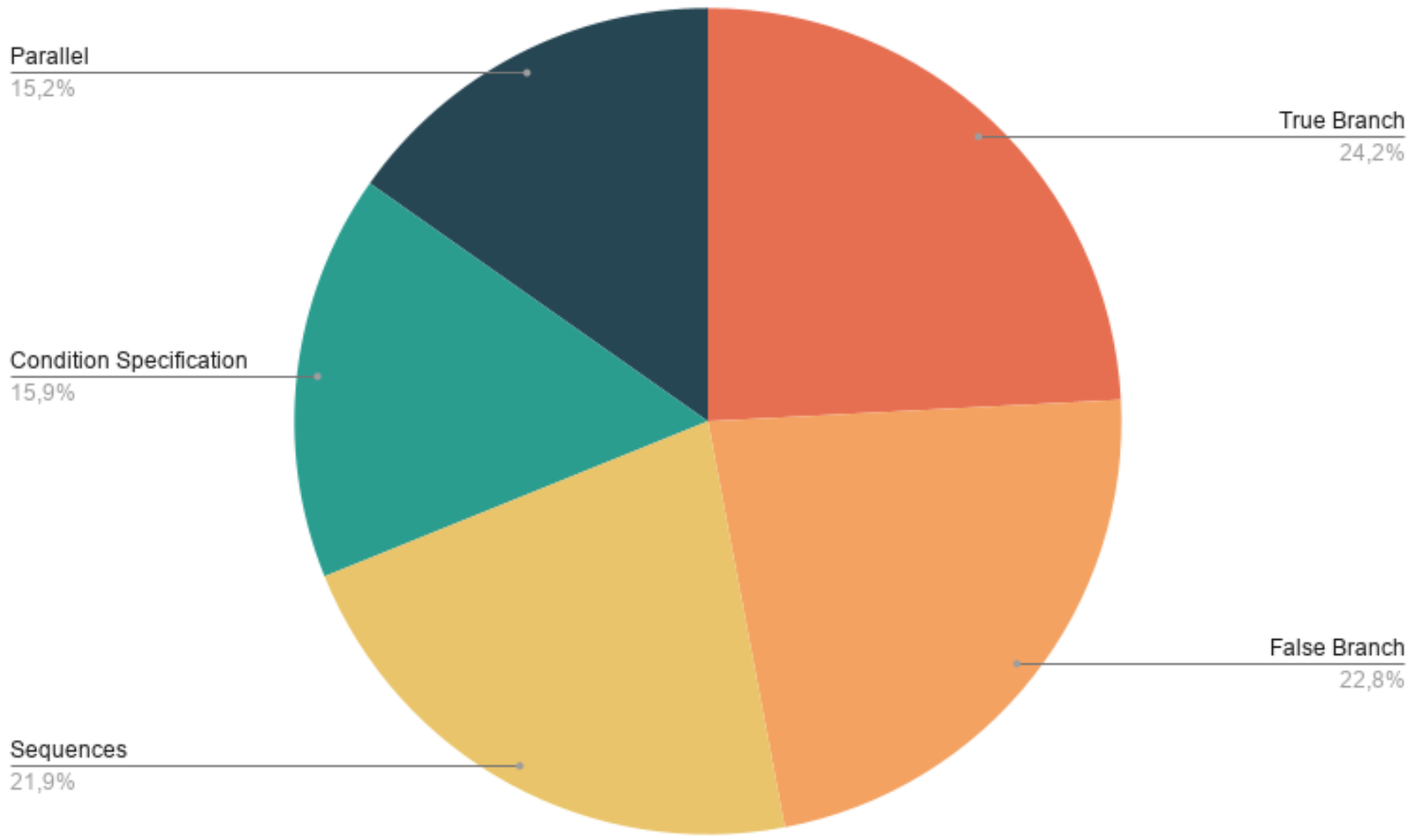

**Fig. 4** Class distribution of the LLM Relation Extraction Evaluation Dataset.

The performance of the LLM-based Relation Extraction is summarized in Table 3 and achieves an overall accuracy of 90.7% and a weighted F1 of 0.91. The proposed control-flow engine manages consistently high performance across all relationship types, particularly for the branching or concurrent ones, demonstrating its ability to handle more complex control-flow structures.

**Table 3** Performance Metrics of the LLM-based Relation Extraction

| ***Relation Class*** | ***Precision*** | ***Recall*** | ***F1*** |
|---|---|---|---|
| $SEQUENCE$ | 1.00 | 0.94 | 0.97 |
| $TRUE_{BRANCH}$ | 1.00 | 0.70 | 0.82 |
| $FALSE_{BRANCH}$ | 0.92 | 1.00 | 0.96 |
| $CONDITION_{SPECIFICATION}$ | 0.77 | 0.96 | 0.85 |
| $PARALLEL$ | 1.00 | 0.91 | 0.95 |

As presented in Figure 5, the proposed solution effectively addresses the $FALSE_{BRANCH}$, $CONDITION_{SPECIFICATION}$ and $PARALLEL$ relations while maintaining high performance on SEQUENCE flows. These results demonstrate that the combination of rule-based validations and large language model relationship inference significantly improves the identification of control-flow relationships, mostly for non-adjacent (non-linear) relations.

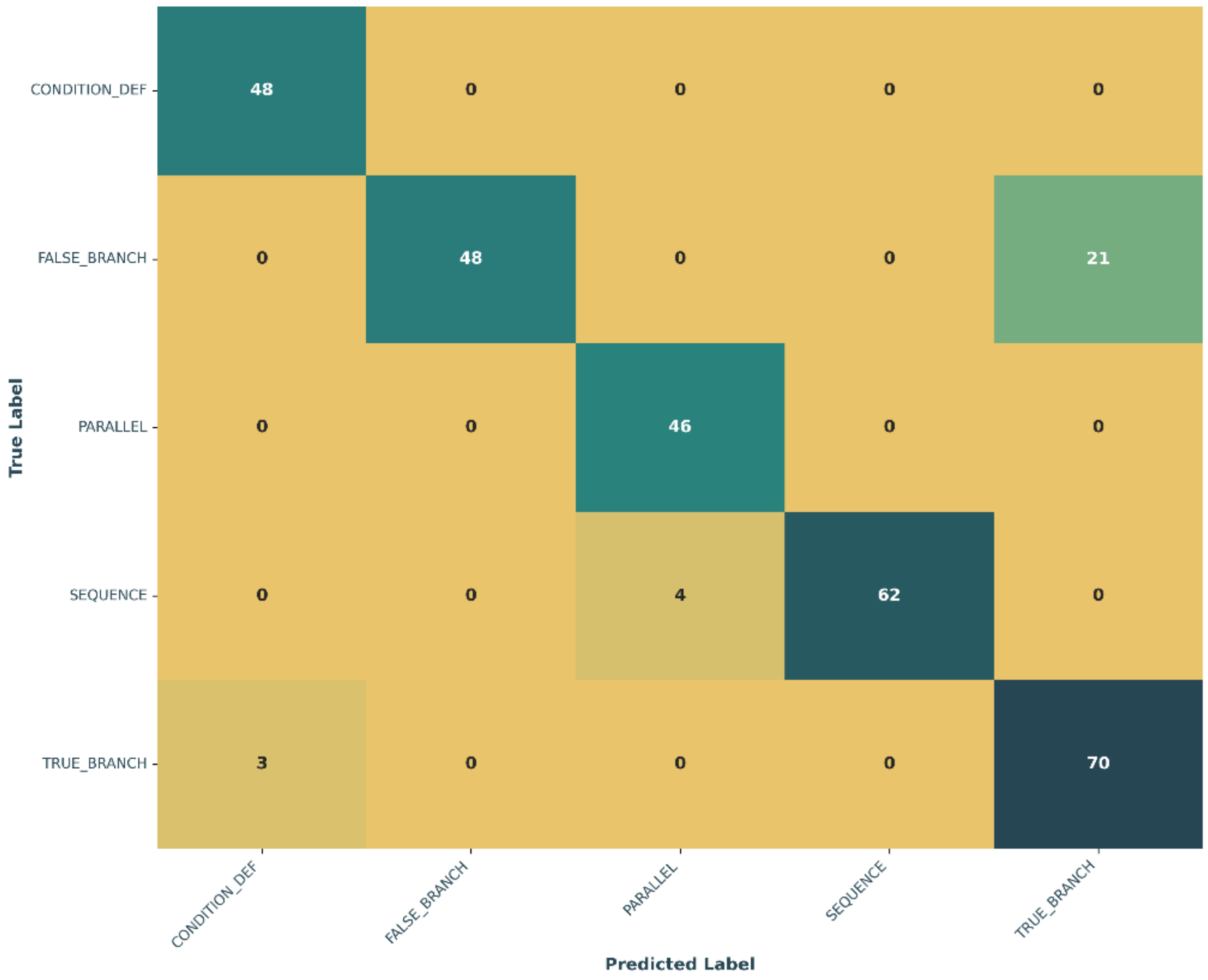

**Fig. 5** Confusion Matrix of the LLM-based Relation Extraction.

To isolate the performance gains strictly to our optimization strategy, we conducted an additional analysis comparing the zero-shot performance of the large language model against our DSPy pipeline. When evaluated in a zero-shot configuration, the LLM achieved an accuracy of around 76%, frequently struggling to distinguish complex conditional branches from standard sequence flows. After applying the optimization pipeline, which dynamically selects the most relevant examples based on contextual similarity, we observed a 14.7% performance gain. This clearly proves that dynamic and programmatically optimized In-Context Learning is necessary in overcoming the hallucinations commonly observed in the zero-shot generative inference. In particular, the optimized prompting strategy reduces hallucinations errors and allows more accurate modelling of complex control-flow dependencies.

## 4.3 Real-world expert defined healthcare process generation

The generation of processes from raw text has multiple applications in healthcare, ranging from the development and updating of treatment plans to the creation of diagnostic and triage protocols, and the organization of workflows within medical teams. Medicine relies on protocols and is in a constant state of evolution, therefore, the continuous updating of these protocols is essential. The ability to generate and update protocols automatically represents a major advantage, saving time and improving overall efficiency. In the prehospital level, simple but continuously improved protocols enable accurate patient triage by facilitating the rapid identification of life-threatening conditions based on presenting symptoms or vital parameters. Once a diagnosis is established, the initial steps: clinical decision-making and treatment, are supported by clearly defined protocols. At the hospital level, monitoring patient parameters according to simplified protocols allows nursing staff to effectively track patient status, recognize early warning signs, and guide the initial response in emergency situations until a physician becomes available. Following discharge, the development of protocols designed for nursing staff or even non-medical personnel may significantly contribute to the monitoring of patients with chronic conditions, particularly in areas with physician shortages or among populations with limited access to medical services.

In this context, to further validate the practical applicability of the proposed solution, we applied the pipeline to a healthcare process model derived from an ongoing European research project, [41]. The process represents an operational care process describing the transitional care path of a patient with heart disease between home and hospital in Romania, reflecting real constraints and multi-actor collaboration patterns. The process description was developed by domain experts and serves as a representative realistic scenario and input for the proposed method.

The process begins when a patient develops new symptoms or experiences a deterioration of a pre-existing condition and depending on the severity of the symptoms, the process branches into two pathways.

If it presents acute severe symptoms, the patient or their family contacts emergency services via 112 or the patient may present directly to the emergency department, either independently or accompanied by family. Upon arrival, the patient is evaluated to determine whether urgent hospitalization is necessary or if they can be discharged with appropriate treatment recommendations.

If discharged, follow-up care is provided by the general practitioner, or the patient may be referred to for outpatient cardiology evaluation.

For mild symptoms or known conditions, the patient visits the general practitioner, who adjusts the current treatment or initiates a new one, and recommends evaluation by a cardiologist. Patients may also consult a cardiologist directly, but only in a private setting.

The outpatient cardiologist performs a detailed evaluation, establishes the final diagnosis, prescribes treatment, and schedules periodic follow-up visits.

For complex cases, the cardiologist recommends hospitalization for additional comprehensive evaluations. If the patient is unable to provide full medical information, the family provides the relevant medical history, otherwise the patient provides it directly.

Based on the final diagnosis, the patient receives the appropriate treatment: medical, interventional, or surgical. During hospitalization, communication with the patient's family is maintained, and they are continuously informed of the patient's progress. When the treatment is complete, the hospital staff prepares the discharge.

At discharge, the hospital staff provides the patient with a discharge summary containing treatment recommendations and a follow-up plan. The process concludes when the patient submits the discharge summary to their general practitioner.

From this description, six actors are identified, nine XOR Gateways and a Parallel path. Patient, Family, Emergency Department, General Practitioner, Outpatient Cardiologist and Hospital represent the actors, serving as primary lanes within the process pool. Following the onset symptoms, the first XOR Gateway evaluates the Patient's condition by distinguishing between acute and mild cases, ensuring a mutually exclusive branching structure. If the symptoms have a sudden onset, the process is directed toward emergency care, either through emergency services (112) being directly contacted by the Patient or Family, or direct presentation to the Emergency Department. Otherwise, the Patient follows the mild pathway and is directed to the General Practitioner or outpatient care. Upon evaluation in the Emergency department, another XOR Gateway determines whether the Patient must be admitted or can be discharged. In the outpatient path, a further XOR Gateway evaluates whether the case is complex or not. If not, the Patient continues with the normal routine care, otherwise, it is referred for urgent hospitalization. Upon admission, another XOR Gateway checks the Patient's ability to provide medical history. If it cannot communicate, the Family provides the medical history, otherwise, the Patient provides it directly and a final diagnosis is established. Finally, a parallel branch is executed during the hospital stay, where treatment (represented as the final XOR Gateway, which can be medical, interventional or surgical) is carried out while communication with Family is maintained. The process ends when the Patient presents the summary to the General Practitioner. This process has been manually translated into a standard BPMN 2.0 diagram (illustrated in Figure 6).

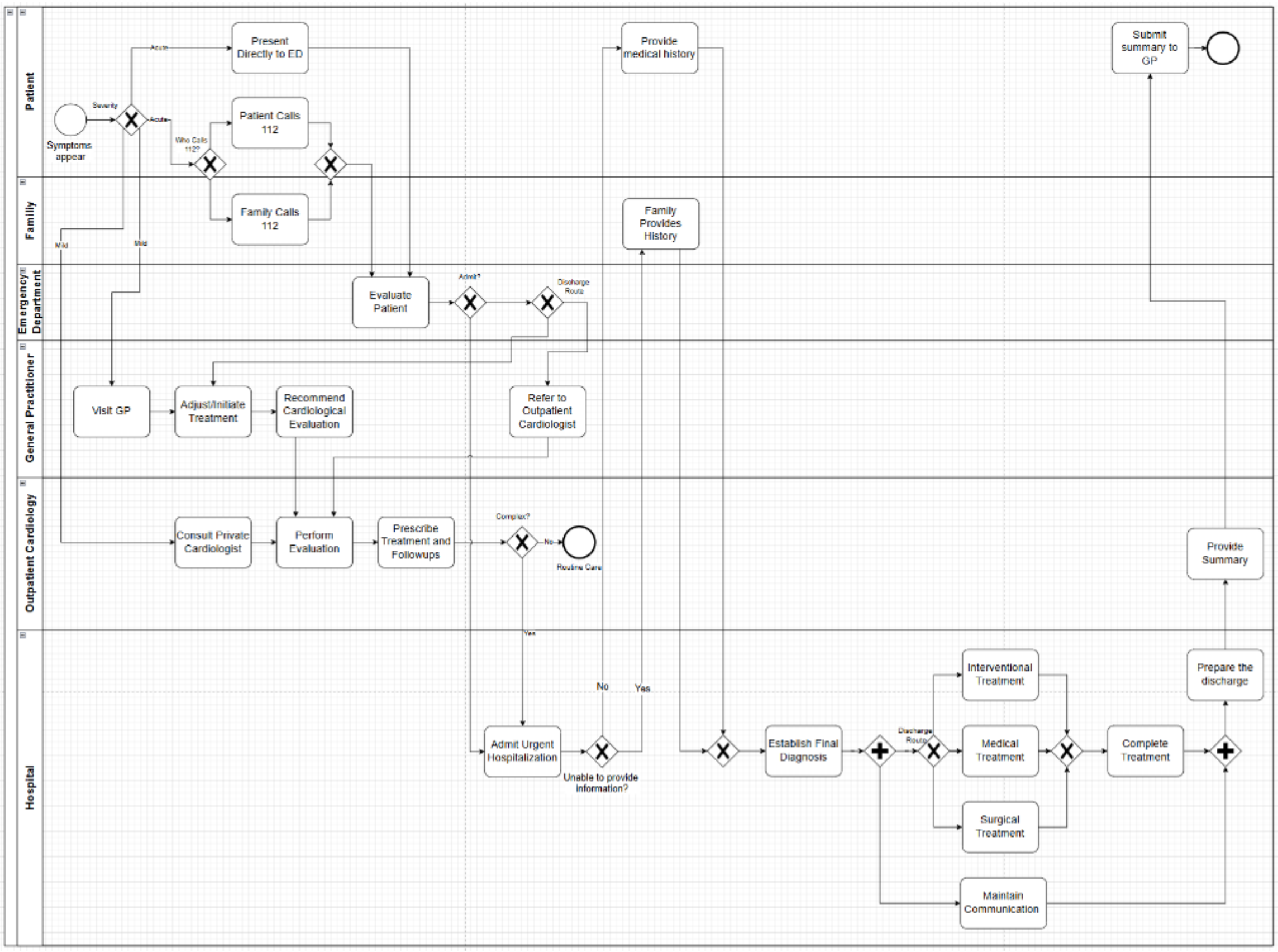

**Fig. 6** Reference BPMN 2.0 Model for healthcare process.

The diagram illustrated in Figure 7 represents the output of our proposed method that successfully identifies the five actors of this process and most of its structure, including the main sequence of activities and decision points. Compared with the original BPMN hand designed process, most of the activities are identified correctly and are aligned with the reference model. However, certain activities that represent intermediary steps and a parallel flow relationship are either simplified or partially omitted, mostly in areas involving implicit conditions. In one instance, when the process starts, an additional activity is introduced, that represents the beginning of the process, which in the reference model is noted as the start event.

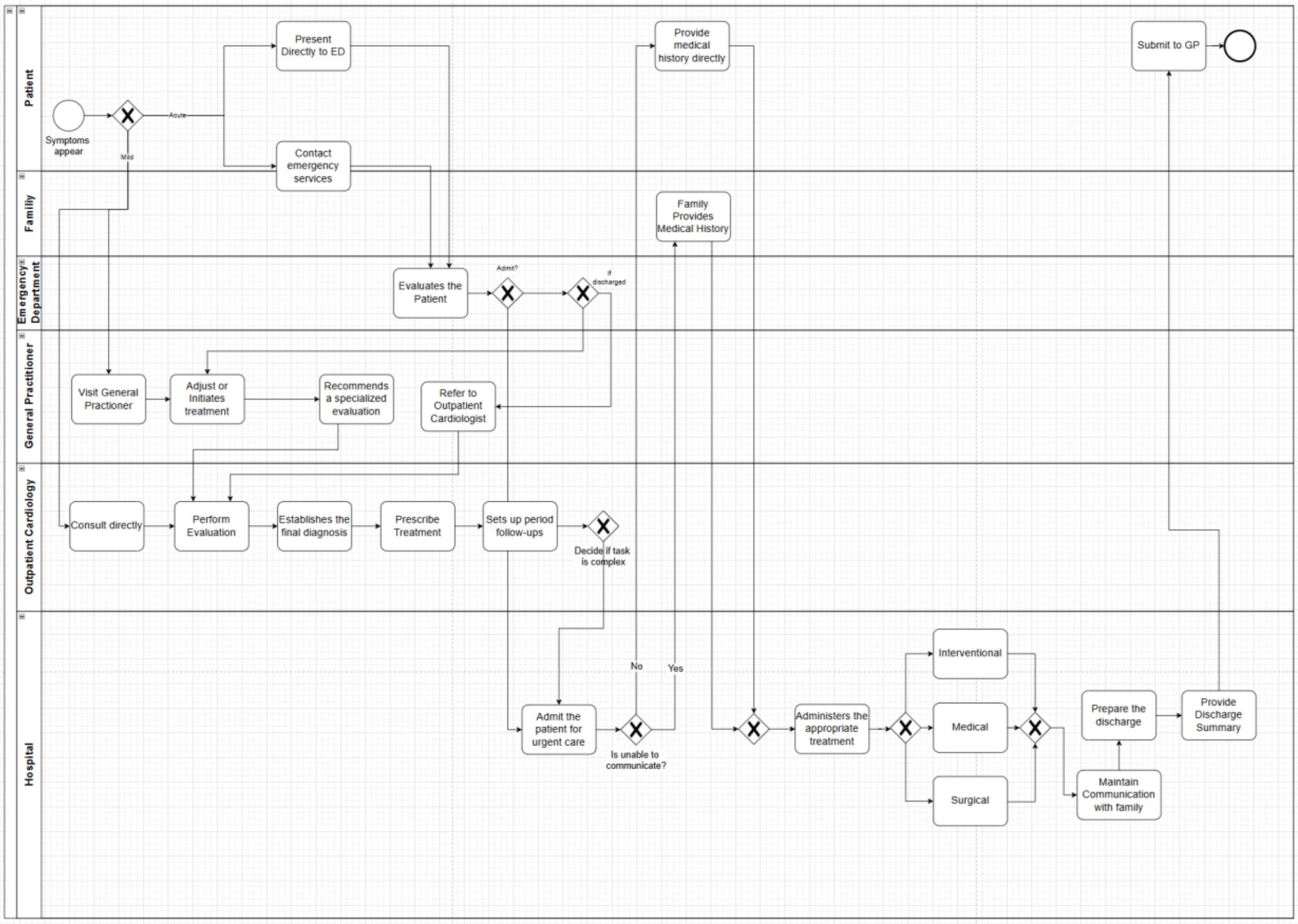

**Fig. 7** Generated BPMN 2.0 Model for healthcare process.

**Table 4** Structural comparison for the healthcare process

| *Element* | *True Positive* | *False Positive* | *False Negative* | *Precision* | *Recall* | *F1* |
|---|---|---|---|---|---|---|
| Tasks | 21 | 1 | 1 | 0.955 | 0.955 | 0.955 |
| Gateways | 6 | 1 | 3 | 0.857 | 0.667 | 0.750 |
| Edges | 19 | 7 | 4 | 0.731 | 0.826 | 0.776 |

Looking at the data presented in Table 4, it indicates that most of the activities have been successfully identified, with one activity missing, an additional one. For the *Prescribe Treatment and Follow-up* activity, the generated model modelled as two distinct activities, which is depicted as only one in the reference process. This showcases that the process can extract information from the description with minimal loss of information. Gateway detection achieves perfect precision, but recall is lower due to missing specific gateways and affects the performance as well of the edge placement in the graph. The *emergency contact* was recognized as an activity performed by both Patient and Family, which is considered on par with reference model even if they are represented differently. Despite these limitations, the overall process maintains the main execution paths, identifies correctly all the participating actors and their interactions, and correctly represents the outcome of decision points. The differences between the two diagrams are mainly structural rather than functional, meaning that the generated model remains valid from a process perspective. This use case highlights the ability of the proposed architecture to generalize complex scenarios, that were not present or have similar representations in the training dataset, while preserving a balance between structural accuracy and correctness.

In addition to structural evaluation, the generated healthcare process was instantiated within the BPMN ontology (illustrated in Figure 8). Each identified entity was mapped to its corresponding ontological class. Activities such as performing initial evaluation, treatment and establishing diagnostic are instantiated as Activity individuals, the patient, family, outpatient cardiologist, general practitioner, emergency department and the hospital staff as Actors, while the inferred control-flow relationships were represented as exclusive Gateway instances. The ontology representation supports further understanding and reutilization of the generated process, and allowing for operations such querying relationships, validating structural constraints, or comparing similar processes. In such processes, it can be helpful, as it allows the reuse of accumulated process knowledge and can facilitate the improvement and customization of similar processes over time.

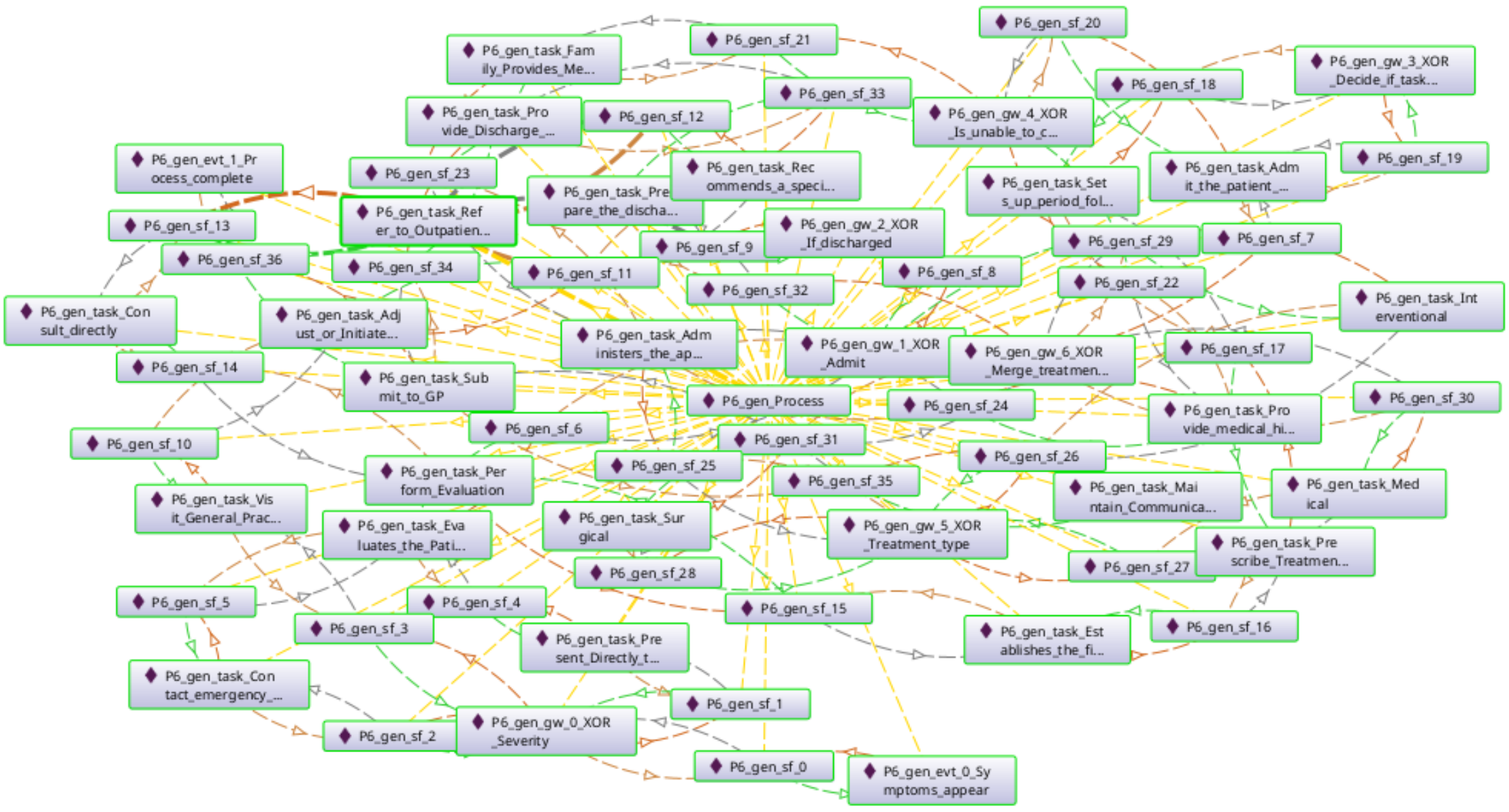

**Fig. 8** BPMN ontology populated with the healthcare process elements.

### *4.4 Camuda training session process generation*

To evaluate the proposed pipeline on standardized BPMN we applied the system to a Goods Dispatch process derived from the Camunda training sessions repository [42]. This example describes a logistics preparation process involving coordination between administrative staff and warehouse personnel. The following textual language description was used as input for the pipeline.

> If goods shall be shipped, the secretary clarifies who will do the shipping. In parallel, the goods can be already packaged by the warehousemen. If you have large amounts, special shipping will be necessary. In these cases the secretary invites three logistic companies to make offers and she selects one of them. In case of small amounts, normal post shipment is used. In the same time, a package label is written by the secretary, and a parcel insurance is taken by the logistics department head if necessary. If everything is ready and merged, the packaged goods are prepared for being picked up by the logistic company.

From the description, several BPMN structural elements can be identified. The process involves three actors: the Secretary, the Warehousemen and the Logistics Department Head. The process begins with a preparation stage, in which shipping responsibilities are clarified while the goods are packaged. A decision point occurs depending on the shipment volume, determining whether the process follows a special shipping procedure involving external logistics companies or standard shipping. Additional tasks are executed in parallel, including the preparation of the package label and the arrangement of parcel insurance. After these steps are done, the goods are prepared for pickup by the selected logistics provider.

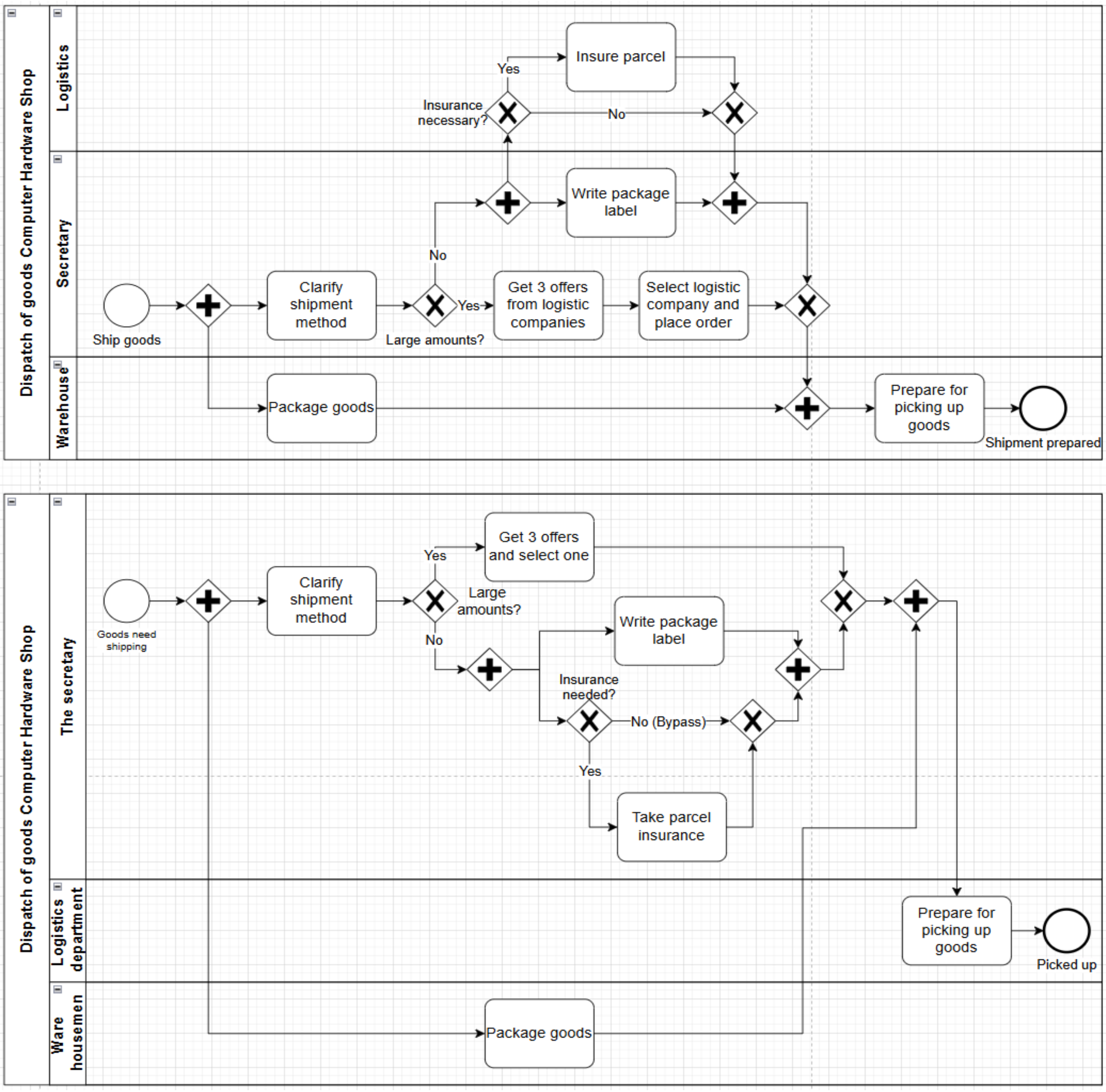

**Fig. 9** Comparison between the reference and the generated BPMN model for Goods Dispatch process

The generated BPMN model, depicted in the lower section of Figure 9, successfully reconstructs most of the process structure presented. The pipeline correctly identifies shipping decision logic and the parallel preparation steps necessary before the final pickup activity. All three participant actors are recognized successfully in the generated model. The pipeline also detects correctly the merger point where the parallel branches converge before the shipment is finished. A minor structural difference appears in the supplier selection stage. Compared to the reference diagram, where the process includes two sequential activities: *Get three offers from logistic companies* and *Select logistic company and place order*, the generated model combined them into a single activity: Get three offers and select one. Although the granularity differs slightly, the semantic meaning of the process follows the described process. Given the activity generated is a merger between the two in reference, it was considered a correct identification.

**Table 5** Structural comparison for the Goods Dispatch process

| *Element* | *True Positive* | *False Positive* | *False Negative* | *Precision* | *Recall* | *F1* |
|---|---|---|---|---|---|---|
| Tasks | 6 | 0 | 0 | 1.000 | 1.000 | 1.000 |
| Gateways | 7 | 1 | 1 | 0.875 | 0.875 | 0.875 |
| Edges | 5 | 1 | 2 | 0.833 | 0.714 | 0.769 |

The structural evaluation presented in Table 5 shows perfect activity detection, indicating that the entity identification layer successfully captures all relevant process activities from the natural language process description. The lower scores in the gateway and edge scores are mainly caused by differences in how branching structures are presented in the generated diagram, mostly due to the merge of the two sequential activities, which affects the number of control-flow relations without affecting the overall process semantics. The results demonstrate that the generated diagram aligns closely with reference one and the textual description.

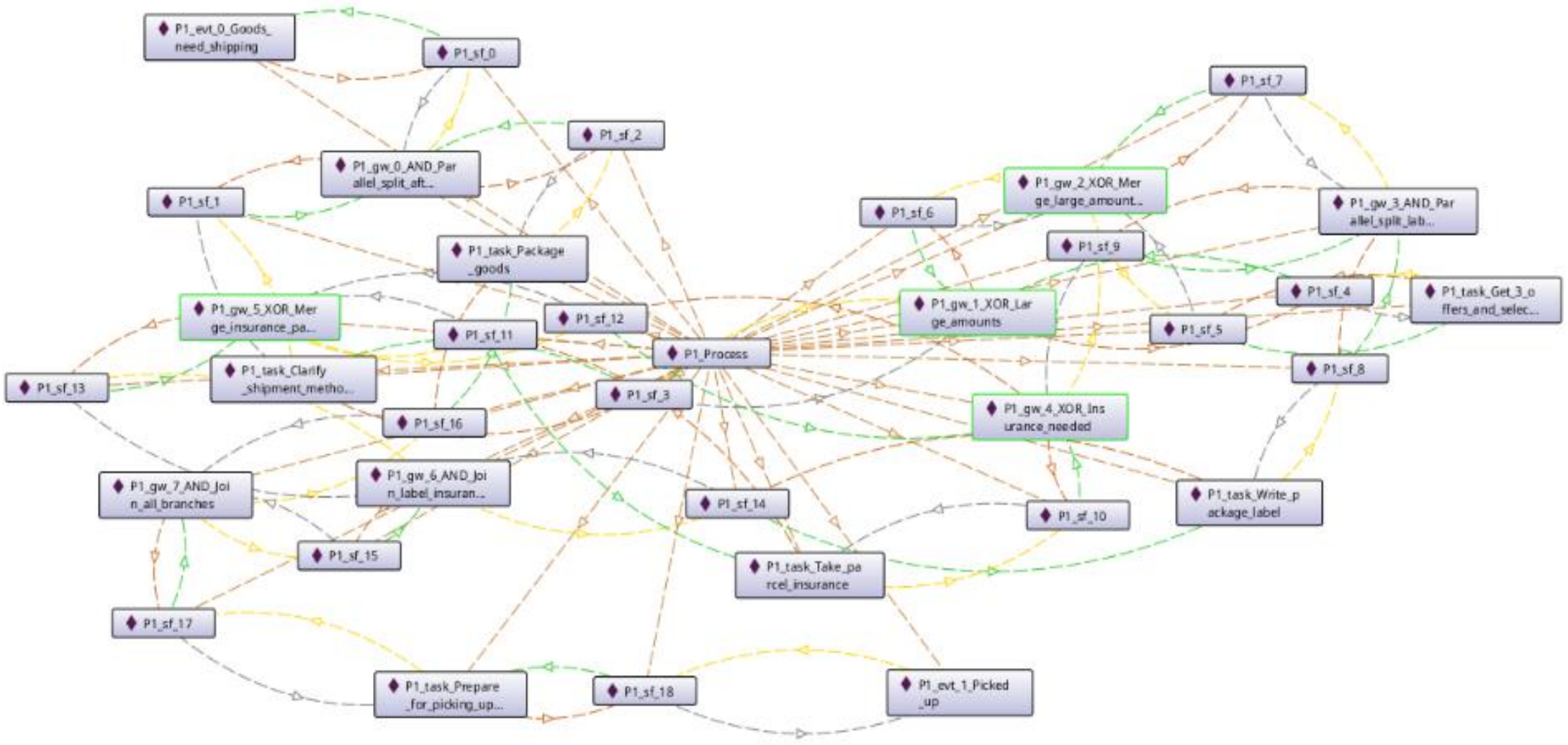

**Fig. 10** BPMN ontology populated for Goods Dispatch process.

The generated process model for the Dispatch of Goods process is also represented within the BPMN ontology (illustrated in Figure 10), where each extracted entity is instantiated as an ontological individual. Activities such as shipment preparation, offer collection, and packaging are represented as instances of Activity class, while secretary, logistic person and warehouse worker are modelled as Actor instances. Ontology captures the control-flow structure through direct relationships between the activities. Conditional branching is represented by the Exclusive Gateway branch, while parallel execution is represented by relationships linking concurrent tasks to a synchronization point.

### *4.5 Textbook business process generation*

Another example evaluated is a Loan Application Eligibility Assessment process from a fundamental business process textbook [43], used as well in financial service scenarios. This process represents the evaluation stage of a loan application in which the applicant's credit history and property appraisal are assessed before a final decision is made.

> A loan application is approved if it passes two checks: (i) the applicant's loan risk assessment, done automatically by a system, and (ii) the appraisal of the property for which the loan has been asked, carried out by a property appraiser. The risk assessment requires a credit history check on the applicant, which is performed by a financial officer. Once both the loan risk assessment and the property appraisal have been performed, a loan officer can assess the applicant's eligibility. If the applicant is not eligible, the application is rejected, otherwise the acceptance pack is prepared and sent to the applicant.

From the textual description, several actors can be identified, including the Loan Officer, the Property Appraiser and an automated Risk Assessment System. However, the reference BPMN model represents only the internal process of the Loan Officer, while the activities performed by other participants are treated as external tasks contributing to the decision-making process. The process contains a parallel evaluation stage in which the applicant's credit history is verified, and the property's appraisal is performed. Once these evaluations are completed, the Loan Officer assesses the applicant's eligibility, resulting either in application rejection or in the preparation of the acceptance package.

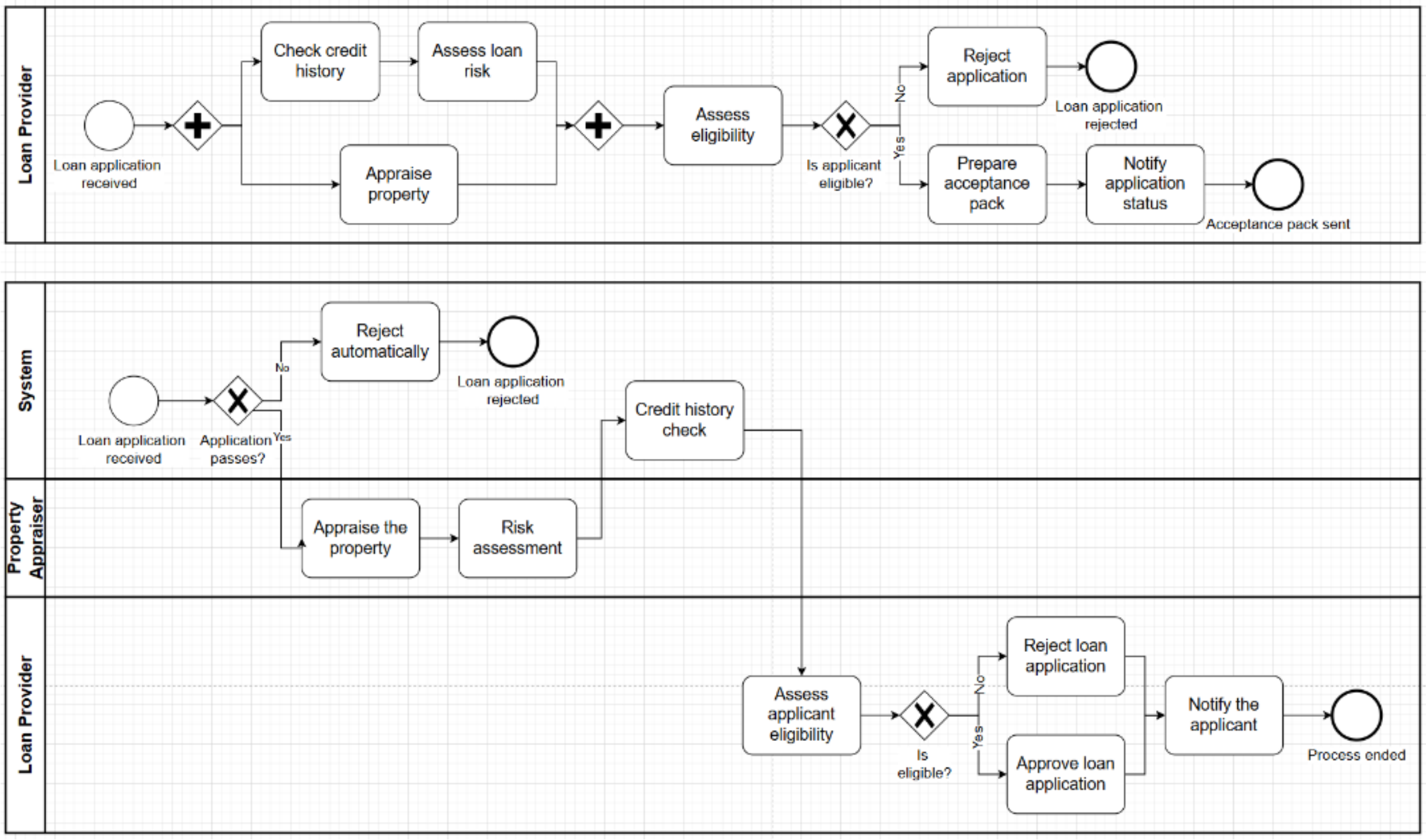

**Fig.** 11 Comparison between the reference and the generated BPMN model for Loan Application Eligibility Assessment process.

The output diagram, presented in Figure 11, identifies correctly all the key activities involved in the eligibility evaluation process. One of the main differences between the generated and reference diagram is the addition of two new actors. In the reference diagram, the actions performed by these two are considered as external information. Thus, the intermediary steps are not specified in the diagram. The system correctly reconstructs the eligibility decision and the approval and rejection paths. However, minor variations occur in the representation of rejection-related activities. The generated model introduces an additional rejection step that splits the rejection outcome into two different tasks in the process. Table 6 shows that the activities were successfully identified, the edge performance is affected due to the additional steps related to rejection and intermediary steps, which subsequently add extra control-flow relations that are not in the reference diagram. The generated model captures the fundamental process logic, including the eligibility decision and parallel verification tasks.

**Table** 6 Structural comparison of the Loan Application Eligibility Assessment process

| *Element* | *True Positive* | *False Positive* | *False Negative* | *Precision* | *Recall* | *F1* |
|---|---|---|---|---|---|---|
| Tasks | 7 | 0 | 0 | 1.000 | 1.000 | 1.000 |
| Gateways | 4 | 1 | 0 | 0.800 | 1.000 | 0.889 |
| Edges | 4 | 3 | 2 | 0.571 | 0.667 | 0.615 |

In the Loan Eligibility process, the ontology representation, illustrated in Figure 12, shows a more intricate process structure involving both parallel verification and conditional decision logic. Activities such as credit history checking, property appraisal, and risk assessment are instantiated as Activity individuals, the system, property appraisal and loan provider as Actors, while the decision point determining eligibility is represented as a Gateway instance. Additionally, the ontology representation includes multiple rejection activities, reflecting the finer granularity introduced by the generated model.

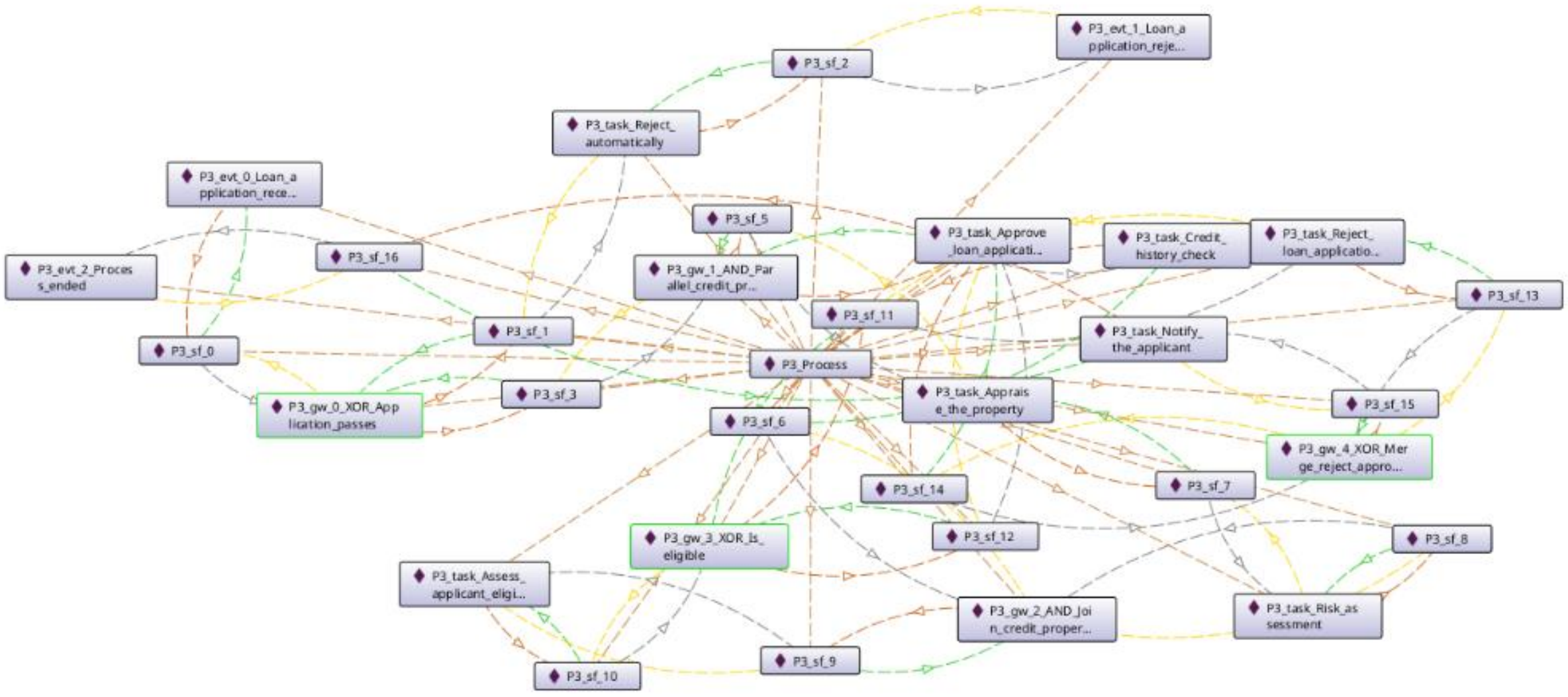

**Fig. 12** BPMN ontology populated for Loan Application Eligibility process.

## 5 Discussions

The proposed solution presents strong potential for automating business process modelling. Enabling direct transformation from unstructured natural language process descriptions into BPMN models, the pipeline reduces the need for specialized modelling expertise, domain-specific knowledge, and accelerates the process design development. This is valuable considering that in organizational environments process knowledge is often documented informally and needs to be manually interpreted. In addition to automation, this approach contributes to improved consistency across process representations. The integration of ontology further extends the applicability of this solution beyond model generation. Transforming process models into knowledge structures allows for further querying and reuse of process knowledge. These capabilities facilitate comparison, detection of recurring patterns, and incremental refinement over time, particularly in organizational environments where processes evolve over time. Evaluation results demonstrate that the proposed multi-layered pipeline can perform relation extraction and structural reconstruction with 90%+ accuracy. A closer analysis of the class level performance shows consistently high precision and recall across all relation categories, confirming the ability to capture both sequential and non-adjacent control-flow structures. This is particularly relevant for modelling conditional and branching structures, where relations span across multiple clauses. The inference engine captures non-adjacent (non-linear) relationships, essential for capturing conditional branching. The performance analysis indicates that structured prompting and controlled inference significantly improves the relation extraction performance and has an important role in reducing misclassification of complex branching patterns.

Another important observation is the ability of the proposed pipeline to handle real world expert defined processes descriptions even under limited data conditions. Despite using a relatively small number of samples, the model achieves high accuracy across the evaluated relation types. This indicates that the structured formulation of the task is composed of a dynamic few-shot conditioning strategy, combining static and dynamic prompt construction, reduces the dependence on large datasets and suggests potential to generalize from limited samples.

At the structural level, the generated BPMN process models have a high degree of similarity to the reference ones across all the evaluated use cases. To assess this similarity, we used the Graph Edit Distance (GED) metric, defined as the cost of the least expensive sequence of edit operations required to transform one graph into another [44]. Formally, this can be expressed as:

$$GED(G_1, G_2) = \min_{\gamma \in \Gamma} \sum_{e \in \gamma} cost(e) \tag{11}$$

where $\gamma$ represents a sequence of edit operations and cost(e) denotes the total cost associated with each operation. Since GED depends on the size of the graphs compared, direct comparison across models of different complexity is not always meaningful. To address this limitation, we use a normalized variant of GED, referred to as Relative Graph Edit Distance (RGED), defined as:

$$RGED(G_1, G_2) = \frac{GED(G_1, G_2)}{\max(|G_1|, |G_2|)} \tag{12}$$

where $|G_1|$ denotes the size of the graph in terms of nodes and edges. Based on the normalized distances, we can further derive a similarity score in the range of [0, 1]:

$$Sim(G_1, G_2) = 1 - RGED(G_1, G_2) \tag{12}$$

where a score of 1 represents identical process models, while lower values indicate greater structural differences.

The Graph Edit Distance (GED) of the healthcare process has a value of 25.41, which represents the total number of modifications required to fully align the generated model with the reference model. This value is highly affected by the absence of the parallel gateway and exclusive gateways, which were chosen in the original model to simplify the initial split into 4 pathways, and the omission of some intermediate activities, rather than due to a failure to reconstruct the process itself. The normalized edited distance (RGED) is 0.438 and the similarity score 0.562. Given the complexity of a healthcare process, that incorporates multiple actors, and actor specific responsibilities, accurately capturing all control-flow relationships is particularly difficult. Although the graph similarity score reflects these discrepancies, it nevertheless suggests that the model produces a closely aligned representation. For the Camunda training process, the computed GED value is 6.87 and the normalized edited distance (RGED) is 0.286 and a similarity score of 0.714. The relatively low edit distances indicate that only a small number of transformations are required to convert to the reference diagram. Most of these differences appear due to variations in task or minor gateway placement adjustments, rather than incorrect control-flow. For the textbook process, GED was 9.97, closely aligned to the previous cases while RGED has a value of 0.453 and the similarity of 0.547. The major structural differences come from variations in the rejection-related activities and additional flow connections introduced in the generated model. Despite these inconsistencies, the main evaluation logic and parallel verification structure remains correctly represented. This analysis shows that, generally, for simpler processes, the generated models align closely with the reference models. GED values are directly proportional to the complexity of the process, due to the additional elements and branching patterns. Nevertheless, the generated models consistently correctly identify the core process topology, including the actors, main execution paths, and decision points.

Despite the promising results, several limitations must be considered. First, the evaluation is partially based on synthetic and generated datasets from relation extraction, due to the absence of publicly available datasets or benchmarks covering BPMN control-flow relationships. While these datasets enable controlled experimentation, they may not fully capture the variability and ambiguity of real process descriptions, potentially leading to an overestimation of the overall performance. Second, although the proposed pipeline shows improvements into non-adjacent control-flow relationships, certain complex linguistic constructs as implicit temporal relation, deeply nested conditions or domain specific terminology, remain challenging and can lead to oversimplification in the generated models. Third, the inference mechanism depends on structured templates and example selection, which introduces a sensitivity to the quality and representativeness of the provided examples. As a result, performance may be limited when dealing with diverse or domain specific descriptions. Finally, the current implementation focuses on a subset of BPMN elements, primarily activities, actors and control-flow constructs, and it does not cover the entire spectrum of BPMN ones, such as event types, message flows, and advanced gateway configurations. This limitation may limit the expressiveness of the generated models in complex scenarios.

## 6 Conclusions

In this paper, we introduced a multi-layered pipeline for transforming natural language process descriptions into BPMN process models. The proposed solution separates the textual processing, control-flow relationship extraction, and structural validation into multiple components, enabling a consistent reconstruction of the processes. A key contribution of this work is the identification of the limitation in adjacent relation extraction approaches and the introduction of an LLM-based inference engine capable of capturing non-adjacent (non-linear) relationships. The experimental evaluation on multiple scenarios, including multi-actor complex processes, demonstrated the practical usage of the proposed method. Beyond model generation, the integration of ontology population extends further the contribution, by providing a foundation for iterative improvement and knowledge reuse. Overall, the proposed pipeline contributes to the automation of business process modelling across diverse domains and scenarios, by bridging the gap between informal process description and formal process representations, while improving consistency and efficiency in process design.

As for future work, several directions can be explored to further improve the proposed approach. First, extending the pipeline to support a broader range of BPMN elements, including event types, message flows, and advance gateways configurations, would increase the level of detail and applicability of the generated models. Second, the ontology representation can be leveraged as a retrieval-augmented generation (RAG) component to support more context-aware inference and improve the handling of complex or domain-specific process descriptions. Finally, integrating learning-based validation mechanisms would help the method to learn, detect and correct structural inconsistencies in generated process models especially in complex or ambiguous scenarios.

**Acknowledgments:** This work was supported by a grant from the Ministry of Research, Innovation and Digitization, CNCS/CCCDI - UEFISCDI, project number ERANET-PARTENERIAT-THCS-TransCare-1, within PNCDI IV.